\documentclass[10pt,twocolumn,letterpaper]{article} 

\usepackage{avss}
\usepackage{times}
\usepackage{epsfig}
\usepackage{graphicx}
\usepackage{amsmath}
\usepackage{amssymb}
\usepackage{diagbox}
\usepackage{caption}
\usepackage{diagbox}
\usepackage{authblk}
\usepackage{subcaption}
\usepackage{adjustbox}
\usepackage{tabularx}

\graphicspath{ {./Images/} }

\usepackage[pagebackref=true,breaklinks=true,letterpaper=true,colorlinks,bookmarks=false]{hyperref}

\def\avssPaperID{30} 

\ifavssfinal\fi
\begin{document}

\title{From Multimodal to Unimodal Attention in Transformers using Knowledge Distillation}



\author[1,2]{Dhruv Agarwal\thanks{These authors contributed equally to this work}\textsuperscript{\dag}}
\author[1]{Tanay Agrawal\textsuperscript{*}\textsuperscript{\ddag}}
\author[1,3]{Laura M. Ferrari\textsuperscript{\ddag}}
\author[1,3]{François Bremond\textsuperscript{\ddag}}

\affil[1]{INRIA Sophia Antipolis - M\'{e}diterran\'{e}e, France}
\affil[2]{Indian Institute of Information Technology, Allahabad, India}
\affil[3]{Université Côte d'Azur, France}
\affil[ ]{\small\textit \textsuperscript{\dag}drv.agwl@gmail.com \textsuperscript{\ddag}name.surname@inria.fr}
\maketitle
\thispagestyle{empty}

\begin{abstract}

Multimodal Deep Learning has garnered much interest, and transformers have triggered novel approaches, thanks to the cross-attention mechanism. Here we propose an approach to deal with two key existing challenges:  the high computational resource demanded and the issue of missing modalities.  We introduce for the first time the concept of knowledge distillation in transformers to use only one modality at inference time.  We report a full study analyzing multiple student-teacher configurations, levels at which distillation is applied, and different methodologies.  With the best configuration, we improved the state-of-the-art accuracy by 3\%, we reduced the number of parameters by 2.5 times and the inference time by 22\%. Such performance-computation tradeoff can be exploited in many applications and we aim at opening a new research area where the deployment of complex models with limited resources is demanded

\end{abstract}

\section{Introduction}

New deep learning models are introduced at a rapid rate for applications in various areas. Among others, transformers
are one of those that have sparked great interest in the computer vision community for vision and multimodal learning \cite{khan2021transformers}. Numerous model variants of the paper "Attention is All You Need" \cite{vaswani2017attention} have been proposed and many works
have centered around the attention mechanism. In multimodal learning the, cross-attention concept has been proven to be an efficient way to incorporate attention in one modality based on others. A key existing challenge is the high computational resource demanded, even at inference time, and therefore the need for GPUs with large memory.

In this paper, we study how to minimize the time and resources required using a transformer based model trained on multimodal data. We utilize the concept of knowledge distillation (KD) to use only one modality at inference time. Basically, KD transfers knowledge from one deep learning model (the teacher) to another (the student), and originally, it was introduced as a technique to reduce the distance between the probability distributions of the output classes of both networks \cite{hinton2015distilling}. When it comes to applying KD to tranformers the first issue is related to the level on which to apply it. 
In a cross-attention transformer the modalities other than the primary one, the query, are  taken as key and value. The primary modality has skip connections allowing it to preserve more information during back-propagation. Thus, when trying to distill information from the teacher (e.g. a cross-attention transformer), at the output level, the student does not have enough knowledge about the other modalities. Thereafter, distillation need to take place at lower feature levels. In order to do so, we propose here the use of Contrastive Representation Distillation (CRD) \cite{tian2020contrastive}, a recent method for KD with state-of-the-art performances which enables transfer of knowledge about representations.
As KD has been effectively applied in deep learning but little has been explored with transformers, a study is here presented comparing multiple student-teacher configurations, KD applied on multiple levels, and comparison of two methods for distillation.

We chose emotion recognition as our application since it is a complex problem where multimodality showed significant improvement \cite{poria2020beneath}. Human emotions possess not only the behavioral component, expressed as visual attributes; they have as well the cognitive and the physiological aspects. This rich information allows us to emphasize and interact with others. For this reason, the use of multimodal inputs, as verbal (e.g. spoken words) and acoustic features (e.g. tone of voice), permitted to combine with the video data complementary salient information increasing the results accuracy.
Nevertheless, the limits of multimodal analysis are various. Besides the high computational cost, one of the main issues is related to missing modalities. 
In the real world, we may not have all the modalities available and some parts of the data can go missing due to hardware, software or human faults. Moreover, the use of heterogeneous data causes synchronization issues, linked to the frequency at which the modalities are recorded.
To address these aspects we propose a novel approach and our contributions are summarized below.
\begin{enumerate}
    \item At the best of our knowledge, we introduce for the first time KD in a transformer architecture for multimodal machine learning. We reduce the computational cost at inference time and able to use just one modality (e.g. video), improving the state-of-the-art accuracy by 3\%.
    \item  We develop a framework to study multiple configurations of student-teacher networks and distillation. We especially compare two methods to distill knowledge, the CRD and the use of cross-entropy loss in the attention module of transformers. We name this second method as Entropy Distillation in Attention Maps (EDAM).
\end{enumerate}
The remaining of this paper is organized as follows. Section 2 reports the main findings in the state-of-the-art with respect to multimodal machine learning, transformers and KD. Section 3 describes the dataset exploited for the experiments. In Section 4, we present the developed detailed framework of the student-teacher network architectures and the KD methods applied. Section 5 discusses the results and section-6 concludes with a summary of our contributions and future perspectives.

\begin{figure*}
\begin{center}
  \epsfig{file=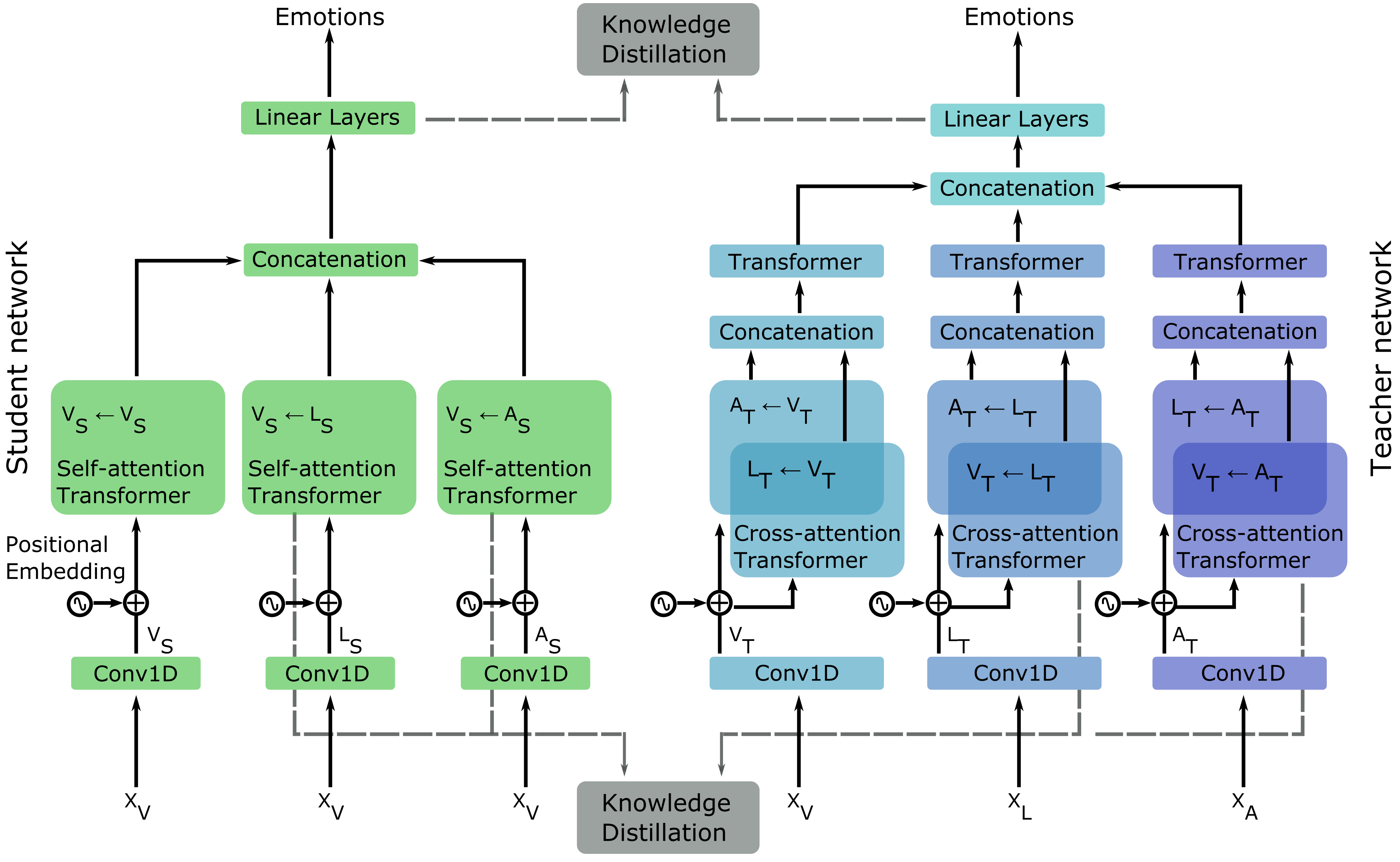, width=5in}
  \caption{The proposed framework, with the student and teacher networks. The knowledge distillation is performed at high level, in the linear layers, and at lower level, inside the transformers. The student network (left) here is the simplified 3-transformers based architecture, while the teacher (right) is the Complete Teacher Network with 9 transformers}
\label{fig:Fig-1}
\end{center}
\end{figure*}
\section{Related Work}
\label{RW}



Multimodal machine learning aims at developing models that can process information from multiple input types. One of the earliest work was in audio-visual speech recognition \cite{yuhas1989integration}, while more recently other fields have started exploiting multimodality as multimedia content indexing \cite{snoek2005multimodal} 
and emotions \cite{busso2008iemocap, zadeh2018multimodal} or personality \cite{nojavanasghari2016deep} recognition. First works in the field proposed the use of early fusion \cite{morency2011towards, perez2013utterance} while late fusion has started to be explored more recently \cite{wang2017select}. 
In the last years more complex architectures have moreover been proposed, as the Tensor Fusion Network for intra- and inter- modality dynamics \cite{Zadeh2017TensorFN} or the use of an attention gate to shift the words through visual and acoustic features \cite{wang2019words}. 
From the famous work by Vaswani  et  al. \cite{vaswani2017attention}, in 2019, Tsai et al. \cite{tsai2019multimodal} proposed the Multimodal Transformer (MulT) to learn representations directly from unaligned multimodal data. In this work cross-modal attention is used to latently adapt elements across modalities, in an end-to-end manner, without explicitly aligning the data and inferring long term dependencies across modalities. In our proposed framework we exploited the MulT architecture for developing the student and teacher network architectures. 


The research field of Multimodal Machine Learning brings some unique challenges, given the heterogeneity of the data, that have been deeply explored \cite{oviatt2018handbook}. One of the main core challenges is the so called co-learning, that is related to the capability of transferring knowledge between modalities. Another core challenge is the representation of heterogeneous data, meaning learning how to summarize multimodal data exploiting their complementarity \cite{baltruvsaitis2018multimodal}. This aspect is linked to the model robustness, that is a relevant theme in real world applications, which in turn is correlated to missing data/modalities. 
Finally, a common general issue of these models is their high computational cost due to their complex architecture, especially when dealing with transformers \cite{kitaev2020reformer}. This is an emerging area of interest and just lately a multimodal end-to-end model have been proposed with a sparse cross-modal attention mechanism to reduce the computational overhead \cite{dai2021multimodal}. An unexplored method to reduce the inference time and dealing with missing modalities is the use of KD. The idea is to leverage on multimodality to build rich representational knowledge that can be transferred to a lighter network, able to infer through just one modality. The concept of KD was firstly introduced by Bucilua et al.~\cite{qwerty12345} and Hinton et al.\cite{ hinton2015distilling}, where a large teacher network is distilled into a small and faster student network to improve the performance of the student at test time. 
From here on a wide variety of works have been proposed \cite{rui2021distill, 
crasto2019mars} and recently many self-supervised methods leveraging on contrastive loss achieved good performance \cite{oord2018representation, chen2020simple}. Among those works, the CRD method proposed by Tian et al. \cite{tian2020contrastive} reported state-of-the-art results proposing to use a contrastive learning strategy for transferring the knowledge in multiple scenarios, comprehending cross-modal transfer and ensemble distillation. 

In this work we analyzed the use of CRD in multiple levels of the proposed framework and in a deeper one, in the attention maps of the transformers, we compared the CRD with what we called the EDAM method. We opted for the EDAM methodology as it is a simpler and runs with less computational cost. This comparison has been performed uniquely at the attention layer as this is the only level that outputs probability distribution, essential for the EDAM loss (i.e. cross-entropy loss).


\section{Dataset}
The dataset used to do experiments in the proposed framework is the CMU-MOSEI dataset \cite{zadeh2018multimodal}, which is one of the largest multimodal datasets for emotion recognition. It is 
made up of 23,454 movie review video clips taken from YouTube. The movie review clips have been selected from more than 1000 speakers and they are gender balanced. The dataset contains three data modalities, video, audio and text. The video and the audio data are directly extracted from the clips while the language data are transcribed and properly punctuated. Each sample from the dataset is labeled by human annotators with continuous values from -3 (strongly negative emotion) to 3 (strongly positive emotion), which we discretized in 7 emotion classes (-3, -2, -1, 0, 1, 2, 3) to model our emotion classification task.
Due to the variable sampling rates of different modalities, there is an inherent non-alignment in the dataset. The dataset is moreover available in both the aligned and non-aligned mode and we used non-aligned one in our task.
We used the extracted features by \cite{tsai2019multimodal} for each modality from the dataset. For video, Facenet \cite{schroff2015facenet} was adopted to give 35 facial action units showing muscle movement for each from the video. For audio, COVAREP \cite{degottex_covarep_2014} is used to get low level acoustic features. As language is the transcript of the audio, the chosen audio features represent the non-semantic features like pitch tracking, loudness and maxima dispersion quotients. For language, Glove word embeddings (glove.840B.300d) \cite{Pennington2014GloveGV} is used to encode the transcript words in 300 dimensional vectors. 
\section{The Proposed Framework}

The framework is a student-teacher network based on transformers. The student network exploits self-attention (uni-modal attention) while the teacher exploits cross-attention (cross-modal attention) on the input sequences, as detailed in section 4.1. The outputs from the transformers are concatenated and passed through linear layers to make the final predictions on emotion classes (Fig~\ref{fig:Fig-1}). In section 4.2 we present the multiple student-teacher configurations experimented here. The capability of transferring knowledge from multiple modalities to one is performed through distillation, at four levels of the framework. As described in section 4.3 we employ KD in the linear layers and in the transformer layers (Fig~\ref{fig:Fig-1}). For each of these cases, we calculate the final loss ($Loss$) as the sum of the distillation loss($L_{KD}$) and classification loss (cross-entropy loss, $L_c$),  as stated in equation (1), which is used to train the student network. While training, we keep the trained weights of the teacher network fixed and only backpropagate gradients through the student network.

    \begin{equation}
       Loss = \alpha L_c + \beta L_{KD}
   \end{equation}
   \hspace{30pt} where $\alpha$, $\beta$ are hyperparameters in the range [0, 1]

\subsection{Teacher and student Transformers}

Teacher and student networks take multimodal sequences (Video, Audio, Language) as input and outputs the likelihood of the 7 emotion classes. The architecture of the networks are derived from \cite{tsai2019multimodal}.
and are detailed below.

\hspace{5pt}

\textbf{Teacher network}: 
The teacher network takes in input 
all the modalities 
and pass them through the respective Conv1D layers, along the temporal dimension, with kernel size of 3. This passage ensures each element in the sequence is aware of its neighboring elements.
The outputs from the Conv1D layers are added with sinusodial positional embeddings before they enter into the transformers, to carry the temporal information \cite{vaswani2017attention}. 
Since we have 3 modalities in the teacher network, we use 6 transformers applying cross-attention in every combination of Query (Q), Key (K) and Value (V) pairs, coming from 2 different modalities. For example, the transformer $V_T\leftarrow A_T$ (Fig~\ref{fig:Fig-1})
represents a transformer where Q is from modality A (Audio) and K and V are from modality V (Video). The transformers $A_T \leftarrow V_T$ and $L_T \leftarrow V_T$ form the \textbf{Video Branch} of the teacher network, as they both use video modality as Q and have the same output dimension at the end of all transformer layers. Similarly, $V_T \leftarrow A_T$ and $L_T \leftarrow A_T$ form the \textbf{Audio Branch}, and $V_T \leftarrow L_T$ and $A_T \leftarrow L_T$ form the \textbf{Language branch}. All three branches together form the \textbf{Complete Teacher Network} represented in Fig~\ref{fig:Fig-1}. 
We experimented on four teacher configurations. One is the Complete Teacher Network and the other three are the individual branches: the Video, the Audio and the Language branches.
The output sequences from each of the 2 transformers in a branch are concatenated along the last dimension to be further input respectively to another set of 3 transformers. This set of transformers applies temporal attention on the fused output of previous transformers. 
The last sequence elements from the output sequences of the 3 transformers are, individually passed (for individual branches), concatenated and passed (for Complete Teacher Network), through linear layers. 

\hspace{5pt}

\textbf{Student Network}: In the student network the Video is the only input modality and it is passed to the model in parallel three times as shown in Fig~\ref{fig:Fig-1}. Each of the inputs goes through a Conv1D layer with different sized kernels to downsample them. Outputs of the $2^{nd}$ and $3^{rd}$ Conv1D layers (from left in Fig~\ref{fig:Fig-1}) are used as proxy for missing audio and language modality sequences and are hence named $A_S$ and $L_S$ respectively. The naming scheme of the transformers in the student network is identical to that of teacher network with subscript "S" in place of "T" to distinguish between "Student" and "Teacher". 

We experimented on four student configurations. One is developed same as the Complete Teacher Network, with 9 transformers. 
The other three configurations are a simplified version of the previous, with 3 transformers. These 3 transformers vary for each student configuration as detailed in the next section, one of those configuration 
(the 5th configuration in section 4.2) is reported in Fig~\ref{fig:Fig-1}.
The last sequence elements of output sequences from all 3 transformers are concatenated and passed through linear layers.

The following section elaborate the different configurations of student-teacher networks.

\subsection{Student-Teacher Configurations}

We study multiple pairs of student and teacher networks to see the effect of distillation on the different architectures.

\begin{enumerate}
    \item \textbf{KD from the Complete Teacher Network}: The teacher is the \textbf{Complete Teacher Network} and the student is constructed exactly the same, except for the inputs. All the 9 cross-attention transformers from the teacher are used for distillation to the corresponding self-attention transformers in the student network.
    

    \item \textbf{KD from Video Branch}: The teacher is the \textbf{Video Branch} and the student network consists of 3 transformers: $V_S \leftarrow V_S$, $A_S \leftarrow V_S$, and $L_S \leftarrow V_S$. The transformers $A_T \leftarrow V_T$ and $L_T \leftarrow V_T$ are used for distillation to corresponding transformers in the student.
   
    \item \textbf{KD from Language Branch}: The teacher is the \textbf{Language Branch} and the student network consists of 3 transformers, $V_S \leftarrow V_S$, $V_S \leftarrow L_S$, and $A_S \leftarrow L_S$. The transformers $V_T \leftarrow L_T$ and $A_T \leftarrow L_T$ are used for distillation to corresponding transformers in the student.
     \item \textbf{KD from Audio Branch}: The teacher is the \textbf{Audio Branch} and the student network consists of 3 transformers, $V_S \leftarrow V_S$, $V_S \leftarrow A_S$, and $L_S \leftarrow A_S$. The transformers $V_T \leftarrow A_T$ and $L_T \leftarrow A_T$ are used for distillation to corresponding transformers in the student.
     
    \item \textbf{KD from Language and Audio Branches of Complete Teacher Network}: The teacher is the \textbf{Complete Teacher Network} and the student network consists of 3 transformers, $V_S \leftarrow V_S$, $V_S \leftarrow L_S$, and $V_S \leftarrow A_S$ as depicted in the Fig~\ref{fig:Fig-1}. The transformers $V_T \leftarrow A_T$ and $V_T \leftarrow L_T$ are used for distillation to corresponding transformers in the student. 
\end{enumerate}

\subsection{Knowledge Distillation}

 

 
 Since different layers in a deep network carry different information, we explore applying distillation to various stages of the network and we compare two methods for loss calculation, CRD and EDAM. We apply CRD at four stages, two at high-level features, in the final and penultimate linear layers, and the other two at the transformer level, where we compare the use of contrastive with cross-entropy loss.
 

\hspace{5pt}

\textbf{Overview of Contrastive Representation Distillation (CRD):}
The CRD method \cite{tian2020contrastive} provides a general framework to bring closer the representation of "positive pairs" from student and teacher networks while pushing apart the representations of "negative pairs". A pair is called positive when the same sample from a dataset is provided to both the networks, while it is negative when different samples from the dataset are input to the networks. 
In the following, we use the expression, \textbf{CRDLoss(X, Y)} to imply CRD being applied on 1D feature vectors X and Y.  



\begin{enumerate}
     \item \textbf{CRD on final and penultimate layers:} CRD is applied on the final and penultimate layers of the student and teacher networks. Here the $L_{KD}$ is calculated by applying the CRD loss at the output of the respective layers from the student and the teacher network, as stated in the following equation
     
     \begin{equation}
         L_{KD} = CRDLoss(Student^{(l)}, Teacher^{(l)})
     \end{equation}
     
     \hspace{20pt} where l indicates the $l^{th}$ layer of the networks.

\item \textbf{CRD on post-attention linear layers:} CRD is applied on the outputs of the attention module of the student-teacher transformer pairs (see Fig-1 in Supplementary Materials). Here the $L_{KD}$ is calculated as stated in the equations (3) - (7) and we take the average of computed $L_{KD}$ for all the pairs when we have multiple pairs in an configuration.

    \begin{equation}
        Attention(Q_X, K_Y, V_Y) = softmax(\dfrac{Q_X K_Y^T}{\sqrt{d_k}}) V_Y
    \end{equation}
    
    \begin{equation}
        DisLvl(Y \leftarrow X) = Attention(Q_X, K_Y, V_Y)
    \end{equation}
    
    where X and Y denote output sequences from Conv1D layers (i.e. $V_S$, $A_S$, $L_S$, $V_T$ $A_T$, and $L_T$), and  Y $\leftarrow$ X indicates that the sequence X serves as Q, while K, V of the transformer come from sequence Y.
    
    Since our transformers have multiple attention layers, we store the $DisLvl(Y \leftarrow X)$ computed from multiple layers in the variable, $DislList$ as follows
    
    \begin{equation}
        DisList^{\alpha_S \leftarrow \beta_S} = [DisLvl(\alpha_S \leftarrow \beta_S)]_{i=1}^{i=l}
    \end{equation}
    
    \begin{equation}
        DisList^{\alpha_T \leftarrow \beta_T} = [DisLvl(\alpha_T \leftarrow \beta_T)]_{i=1}^{i=l}
    \end{equation}
    
    where $\alpha$ and $\beta$ are input modalities (Video, Audio, or Language), subscripts "S" and "T" refer to Student and Teacher respectively, and $l$ is the number of layers in the student and teacher transformers. Finally, $L_{KD}$ is calculated using equation (7)
    
    \small
    \begin{equation}
        L_{KD}^{\alpha \leftarrow \beta} = \dfrac{\sum_{i=1}^{l} CRDLoss(DisList_i^{\alpha_S \leftarrow \beta_S}, DisList_i^{\alpha_T \leftarrow \beta_T})}{l}
    \end{equation}
    \normalsize

    \item \textbf {CRD on Attention Maps:}. CRD loss is applied on the attention map of each transformer pairs (see Fig-1 in supplementary Materials). The attention map of a transformer layer is described in the following equation (8). 
    
    \begin{equation}
        DisLvl(Y \leftarrow X) = softmax(\dfrac{Q_X K_Y^T}{\sqrt{d_k}})
    \end{equation}
    
    \hspace{5pt}

    In our cross-attention transformers (the teacher network), Q comes from different modality than K and V while all are from the same modality in case of self-attention transformers (in student network). The teacher network benefits from multiple modalities as it has the opportunity of extracting features from all combinations of modalities put as Q and (K, V), which however is not the case with student networks. Distilling different attention maps from multiple transformers can account for the missing modalities in the student networks to an extent and can be conducive in learning richer representations. 
    In an attention map, the sum of values in every row is \(1\) and every cell has value in the range \(0-1\). We therefore see every row as a probability distribution over n classes where n is the number of columns in the attention map of dimension $m \times n$. Through distillation we aim to bring closer the probability distribution of student network to the teacher network and at the same time push apart the distributions of positive samples away from negative samples. ($L_{KD}$) for one transformer pair is thus calculated using equations, (5), (6) (with $DisLvl$ defined in equation (8)), and (9). In the configurations with multiple transformer pairs, average over $L_{KD}$ obtained from all pairs serves as the final $L_{KD}$
    
    
    
    \small
    \begin{equation}
    \resizebox{\linewidth}{!}{$L_{KD}^{\alpha \leftarrow \beta} = \dfrac{\sum_{i=1}^{l}\sum_{j=1}^{m} CRDLoss(DisList_{ij}^{\alpha_S \leftarrow \beta_S}, DisList_{ij}^{\alpha_T \leftarrow \beta_T})}{ml}$}
    \end{equation}
    \normalsize
    
    where $DisList_{ij}^{Y \leftarrow X}$ denotes the $j^{th}$ row of the $i^{th}$ layer attention map from transformer $Y \leftarrow X$.
    m = number of rows in attention map and l = number of layers in the transformer

\item \textbf{EDAM on Attention Maps:} The cross-entropy loss is applied on the attention maps of each transformer pairs. For an attention map of m $\times$ n dimension, we view the rows as probability distribution over n classes and define the distillation loss as a modification to classical cross-entropy loss between student and teacher attention map rows. Similar to the case of CRD on attention maps, we obtain attention maps from all transformer layers using equations (5) and (6) (with Attention map / DisLvl defined using equation (10) and (11))


    \begin{equation}
        DisLvl(Y \leftarrow X, t) = temp\_softmax(\dfrac{Q_X K_Y^T}{\sqrt{d_k}}, t)
    \end{equation}
    
    \begin{equation}
        temp\_softmax(X, t)^{(i)} = \dfrac{exp(X^{(i)}/t)} {\sum_{j=1}^{k} exp(X^{(j)} / t)}
    \end{equation}

    where t is the temperature parameter which adjusts the sharpening of distribution.

\hspace{5pt}
    
and calculate the distillation loss for a pair using the equations (12) and (13). 

\small
\begin{equation}
    L_{KD}^{\alpha \leftarrow \beta} = \dfrac{\sum_{i=1}^{l}\sum_{j=1}^{m} F(DisList_{ij}^{\alpha_S \leftarrow \beta_S}, DisList_{ij}^{\alpha_T \leftarrow \beta_T})}{ml}
\end{equation}
\normalsize

\begin{equation}
    F(a, b) = -alog(b)
\end{equation}

For the configuration with distillation on multiple transformer pairs, final $L_{KD}$ is calculated by averaging over $L_{KD}$ obtained from all pairs.

We employ the EDAM in two ways, named \textbf{EDAM-S$\downarrow$} and \textbf{EDAM-T$\uparrow$} to deal with diverse dimensions, due to the different inputs of the teacher and the student. The teacher has multiple modalities as input while the student has only one modality, and each modality may have a different sequence length. Thereafter the attention map in the student and teacher networks can be of different dimensions. To handle this, in \textbf{EDAM-S$\downarrow$} we downsample the Video inputs to the student, $A_S$ and $L_S$ using Conv1D to match them with Audio and Language input in teacher network, $A_T$ and $L_T$ respectively.

Instead of downsampling inputs, in \textbf{EDAM-T$\uparrow$} we upsample the attention maps in the teacher transformer layers to match with the dimensions of attention maps in the student transformer layers using linear layers.

    

\end{enumerate}

\section{Results and Discussion}
Table-\ref{table:nonlin} reports the results obtained with the proposed framework by varying the teacher network design and the KD levels. The table is arranged with alternating student and teacher rows. The first row shows the result for just the student network, without KD. Following that, every teacher-student row pair has the teacher network performance and the performance for student, with different distillation methods (Section-4.3).

\begin{table}[hbt!]

 \centering 
\newcolumntype{b}{X}
\newcolumntype{s}{>{\hsize=.33\hsize}X}

\begin{tabularx}{\columnwidth}{|s|b|s|s|} 

\hline
\emph{Network} & \emph{Description} & \emph{Accuracy (\%)} & \emph{F1-score} \\ [1ex] 
\hline\hline
Student & Without KD & 41.296 & 32.142 \\ [1ex]
\hline
\end{tabularx}

\begin{tabular}{l}
KD from the Complete Teacher Network\\
\end{tabular}

\begin{tabularx}{\columnwidth}{|s|b|s|s|}
\hline
Teacher & Complete Teacher Network & 49.779 & 48.817 \\ [1ex]
\hline
Student & EDAM-S$\downarrow$ on Attention Maps & 44.137 & 36.286 \\
        [1ex]
\hline
\end{tabularx}

\begin{tabular}{l}
KD from Video Branch\\
\end{tabular}

\begin{tabularx}{\columnwidth}{|s|b|s|s|}
\hline
Teacher & Video Branch & 49.621 & 48.542 \\ [1ex]
\hline
Student & CRD on final linear layer & 43.711 & 33.281 \\
        & CRD on penultimate linear layer & 42.576 & 35.565 \\
        & CRD on attention maps & 43.837 & 35.912 \\
        [1ex]
\hline
\end{tabularx}

\begin{tabular}{l}
KD from Language Branch\\
\end{tabular}

\begin{tabularx}{\columnwidth}{|s|b|s|s|}
\hline
Teacher & Language Branch & 49.217 & 47.250 \\ [1ex]
\hline
Student & CRD on final linear layer & 43.154 & 32.863 \\
        & CRD on penultimate linear layer & 42.901 & 34.467 \\[1ex]
\hline
\end{tabularx}

\begin{tabular}{l}
KD from Audio Branch\\
\end{tabular}

\begin{tabularx}{\columnwidth}{|s|b|s|s|}
\hline
Teacher & Audio Branch & 49.217 & 47.232 \\ [1ex]
\hline
Student & CRD on final linear layer & 43.584 & 33.260 \\
        & CRD on penultimate linear layer & 43.232 & 35.115 \\[1ex]
\hline
\end{tabularx}

\begin{tabular}{l}
KD from Language and Audio Branches\\
\end{tabular}

\begin{tabularx}{\columnwidth}{|s|b|s|s|} 
\hline
Teacher & Complete teacher network & 49.779 & 48.817 \\ [1ex]
\hline
Student & CRD on final linear layer & 42.731 & 32.513 \\
        & CRD on penultimate linear layer & 43.732 & 35.632 \\
        & CRD on post-attention linear layers & 43.837 & 35.711 \\
        & CRD on attention maps  & 44.031 & 36.110 \\
        & EDAM-S$\downarrow$ on attention maps & \textbf{44.231} & \textbf{36.332} \\
        & EDAM-T$\uparrow$ on attention maps & 43.993 & 36.189 \\ [1ex]
\hline
\end{tabularx}
\hspace{5pt}
\caption{Results on CMU-MOSEI Dataset - Unaligned}
\label{table:nonlin} 
\end{table}

We achieved the best result with the simplified Student Network and the Complete Teacher Network (the 5th configuration in section 4.2), through the EDAM-S$\downarrow$ method on attention maps. The accuracy in this case is 44.231\% which improves the one-modality case (row-1 of the Table-\ref{table:nonlin}: 41.296 \%), performed with state-of-the-art architecture \cite{tsai2019multimodal}, by approximately 3\% in accuracy and 4 points in F1-score. We further compared the number of parameters of our best performing student network (0.675 Million) to the complete teacher network (1.802 Million). The parameters are reduced of around 2.5 times, which leads to a minimized inference time of 29ms (22\% less than the Complete Teacher Network). Note that, the results obtained for KD from Complete Teacher Network (row-2 in Table-\ref{table:nonlin}) were not optimized rigorously as the behavior of the smaller student (row-5 in Table-\ref{table:nonlin}) was similar to it.

Notably, the improvements, in accuracy and F1-score, occurred in the lower layers of the architecture, in the attention map. In the Fig-3 of Supplementary Material we compare the first two layers of the attention maps of the teacher and of the students, with and without KD. Here we clearly show the improvement got through KD application.
This proves our initial hypothesis, suggesting that higher layers do not provide enough information in cross-attention transformers to the student network. 
Moreover, by comparing CRD with EDAM methods we found out that EDAM, the lighter one, gives the best results. Therefore we were able to obtain the highest accuracy and F1-score with the less computational requirement. 
A challenge with applying distillation to attention maps is that the dimensions of the teacher and student are different as the student capitalizes on self attention, while the teacher on cross-attention. We explored two methodologies to overcome this issue, downsampling (EDAM-S$\downarrow$) and upsampling (EDAM-T$\uparrow$), showing that the former works better. This result is intuitive as the features are designed to be as orthogonal as possible and reducing the map size using a linear layer will lead to loss of information. 

In this work we used unaligned modalities as input. Nevertheless, if aligned modalities are adopted instead, the downsampling method for which loss of information is linked to the reduced temporal resolution, would have shown even better performance.

Note that in Table-\ref{table:nonlin} we did not optimize the hyperparameters for the CRD on attention map, as the outputs were closed to the EDAM method that do not required any additional parameters for training.

Looking at the diverse teacher configurations, the use of the complete teacher network or of the individual teacher branches, does not have a significant impact on the results (see results of teacher networks in the Table-\ref{table:nonlin}). This can be due to the high temporal and semantic correlation among the modalities with respect to the output. On the other hand, individual branches teacher are smaller in size, therefore quicker to train. 

Finally, looking at the results, we noted that in some cases (e.g. in the case of Teacher with Language Branch) accuracy decreases while the F1-score increases when applying KD in deeper layers. This can be due to the fact the data is skewed and the better accuracy reached in the higher layers is obtained by giving the majority class as output most of the times. When applying KD in the deeper layers the network better learns to take into account the other classes, so the accuracy decreases. Controversially the F1 score, which is a good metric for imbalanced data, is aligned with the expected trend, where the results improve applying KD in deeper layers.

\section{Conclusions}

In this work we explore KD at various levels and with multiple student-teachers configurations. We establish that going deeper in the transformer network is conducive to KD by getting best results with KD on attention maps.

In this first work we used as single modality, in the student network, the video; nevertheless, the same methodology can be adopted to other modalities.
Studying the multiple teacher configurations we demonstrated that using different modalities as query, key and value does not have much impact on the results. This will help future studies, to reduce the effort in the evaluation of diverse permutations. 

Since this is an unexplored field, multiple directions are envisioned as future work. First, others techniques of distillation can be explored. The EDAM-S$\downarrow$ method, based on  simple cross-entropy loss achieved the best accuracy and F1-score, and required fewer learning parameters compared to KD with CRD loss. This serves as a proof of concept and opens the possibility for experimenting different loss functions for KD on attention maps. The accuracy of one-modality based models can be moreover improved by exploiting other modalities, such as the physiological data. We believe that, while strong emotions can be better recognized by using one modality with distilled knowledge, subtle emotions are more complicated to be addressed and physiological cues could add relevant dependencies in the final one-modality.
We believe our approach is robust and extending it to other datasets and fields of work is a pertinent next step. 

We aim that this paper will pave the way for a novel research area in multimodal machine learning focusing on resources reduction. This will permit the use of complex algorithms when the inference time has to be minimized, exploiting a richer embedding space and increased accuracy. An example of application is in the field of emotion recognition, where high computational cost can not be sustained and some modalities might be partly missing.

{\small
\bibliographystyle{ieee}
\bibliography{egpaper}

\begin{thebibliography}{10}\itemsep=-1pt

\bibitem{baltruvsaitis2018multimodal}
T.~Baltru{\v{s}}aitis, C.~Ahuja, and L.-P. Morency.
\newblock Multimodal machine learning: A survey and taxonomy.
\newblock {\em IEEE transactions on pattern analysis and machine intelligence},
  41(2):423--443, 2018.

\bibitem{qwerty12345}
C.~Buciluǎ, R.~Caruana, and A.~Niculescu-Mizil.
\newblock Model compression.
\newblock In {\em Proceedings of the 12th ACM SIGKDD international conference
  on Knowledge discovery and data mining}, pages 535--541, 2006.

\bibitem{busso2008iemocap}
C.~Busso, M.~Bulut, C.-C. Lee, A.~Kazemzadeh, E.~Mower, S.~Kim, J.~N. Chang,
  S.~Lee, and S.~S. Narayanan.
\newblock Iemocap: Interactive emotional dyadic motion capture database.
\newblock {\em Language resources and evaluation}, 42(4):335--359, 2008.

\bibitem{chen2020simple}
T.~Chen, S.~Kornblith, M.~Norouzi, and G.~Hinton.
\newblock A simple framework for contrastive learning of visual
  representations, 2020.

\bibitem{crasto2019mars}
N.~Crasto, P.~Weinzaepfel, K.~Alahari, and C.~Schmid.
\newblock Mars: Motion-augmented rgb stream for action recognition.
\newblock In {\em Proceedings of the IEEE/CVF Conference on Computer Vision and
  Pattern Recognition}, pages 7882--7891, 2019.

\bibitem{dai2021multimodal}
W.~Dai, S.~Cahyawijaya, Z.~Liu, and P.~Fung.
\newblock Multimodal end-to-end sparse model for emotion recognition.
\newblock {\em arXiv preprint arXiv:2103.09666}, 2021.

\bibitem{degottex_covarep_2014}
G.~Degottex, J.~Kane, T.~Drugman, T.~Raitio, and S.~Scherer.
\newblock {COVAREP} - {A} collaborative voice analysis repository for speech
  technologies.
\newblock In {\em Proceedings of {IEEE} {International} {Conference} on
  {Acoustics}, {Speech} and {Signal} {Processing} ({ICASSP} 2014)}, pages
  960--964, Florence, Italy, May 2014. IEEE.

\bibitem{hinton2015distilling}
G.~Hinton, O.~Vinyals, and J.~Dean.
\newblock Distilling the knowledge in a neural network, 2015.

\bibitem{khan2021transformers}
S.~Khan, M.~Naseer, M.~Hayat, S.~W. Zamir, F.~S. Khan, and M.~Shah.
\newblock Transformers in vision: A survey.
\newblock {\em arXiv preprint arXiv:2101.01169}, 2021.

\bibitem{kitaev2020reformer}
N.~Kitaev, Łukasz Kaiser, and A.~Levskaya.
\newblock Reformer: The efficient transformer, 2020.

\bibitem{morency2011towards}
L.-P. Morency, R.~Mihalcea, and P.~Doshi.
\newblock Towards multimodal sentiment analysis: Harvesting opinions from the
  web.
\newblock In {\em Proceedings of the 13th international conference on
  multimodal interfaces}, pages 169--176, 2011.

\bibitem{nojavanasghari2016deep}
B.~Nojavanasghari, D.~Gopinath, J.~Koushik, T.~Baltru{\v{s}}aitis, and L.-P.
  Morency.
\newblock Deep multimodal fusion for persuasiveness prediction.
\newblock In {\em Proceedings of the 18th ACM International Conference on
  Multimodal Interaction}, pages 284--288, 2016.

\bibitem{oord2018representation}
A.~v.~d. Oord, Y.~Li, and O.~Vinyals.
\newblock Representation learning with contrastive predictive coding.
\newblock {\em arXiv preprint arXiv:1807.03748}, 2018.

\bibitem{oviatt2018handbook}
S.~Oviatt, B.~Schuller, P.~Cohen, D.~Sonntag, G.~Potamianos, and A.~Kr{\"u}ger.
\newblock {\em The handbook of multimodal-multisensor interfaces, Volume 2:
  Signal processing, architectures, and detection of emotion and cognition}.
\newblock Morgan \& Claypool, 2018.

\bibitem{Pennington2014GloveGV}
J.~Pennington, R.~Socher, and C.~D. Manning.
\newblock Glove: Global vectors for word representation.
\newblock In {\em EMNLP}, 2014.

\bibitem{perez2013utterance}
V.~P{\'e}rez-Rosas, R.~Mihalcea, and L.-P. Morency.
\newblock Utterance-level multimodal sentiment analysis.
\newblock In {\em Proceedings of the 51st Annual Meeting of the Association for
  Computational Linguistics (Volume 1: Long Papers)}, pages 973--982, 2013.

\bibitem{poria2020beneath}
S.~Poria, D.~Hazarika, N.~Majumder, and R.~Mihalcea.
\newblock Beneath the tip of the iceberg: Current challenges and new directions
  in sentiment analysis research.
\newblock {\em IEEE Transactions on Affective Computing}, 2020.

\bibitem{rui2021distill}
F.~B. Rui~Dai, Srijan~Das.
\newblock Learning an augmented rgb representation with cross-modal knowledge
  distillation for action detection. accepted, available at hal-03314575,.
\newblock {\em IEEE/CVF International Conference on Computer Vision (ICCV)}.

\bibitem{schroff2015facenet}
F.~Schroff, D.~Kalenichenko, and J.~Philbin.
\newblock Facenet: A unified embedding for face recognition and clustering.
\newblock In {\em Proceedings of the IEEE conference on computer vision and
  pattern recognition}, pages 815--823, 2015.

\bibitem{snoek2005multimodal}
C.~G. Snoek and M.~Worring.
\newblock Multimodal video indexing: A review of the state-of-the-art.
\newblock {\em Multimedia tools and applications}, 25(1):5--35, 2005.

\bibitem{tian2020contrastive}
Y.~Tian, D.~Krishnan, and P.~Isola.
\newblock Contrastive representation distillation, 2020.

\bibitem{tsai2019multimodal}
Y.-H.~H. Tsai, S.~Bai, P.~P. Liang, J.~Z. Kolter, L.-P. Morency, and
  R.~Salakhutdinov.
\newblock Multimodal transformer for unaligned multimodal language sequences.
\newblock In {\em Proceedings of the conference. Association for Computational
  Linguistics. Meeting}, volume 2019, page 6558. NIH Public Access, 2019.

\bibitem{vaswani2017attention}
A.~Vaswani, N.~Shazeer, N.~Parmar, J.~Uszkoreit, L.~Jones, A.~N. Gomez,
  L.~Kaiser, and I.~Polosukhin.
\newblock Attention is all you need, 2017.

\bibitem{wang2017select}
H.~Wang, A.~Meghawat, L.-P. Morency, and E.~P. Xing.
\newblock Select-additive learning: Improving generalization in multimodal
  sentiment analysis.
\newblock In {\em 2017 IEEE International Conference on Multimedia and Expo
  (ICME)}, pages 949--954. IEEE, 2017.

\bibitem{wang2019words}
Y.~Wang, Y.~Shen, Z.~Liu, P.~P. Liang, A.~Zadeh, and L.-P. Morency.
\newblock Words can shift: Dynamically adjusting word representations using
  nonverbal behaviors.
\newblock In {\em Proceedings of the AAAI Conference on Artificial
  Intelligence}, volume~33, pages 7216--7223, 2019.

\bibitem{yuhas1989integration}
B.~P. Yuhas, M.~H. Goldstein, and T.~J. Sejnowski.
\newblock Integration of acoustic and visual speech signals using neural
  networks.
\newblock {\em IEEE Communications Magazine}, 27(11):65--71, 1989.

\bibitem{Zadeh2017TensorFN}
A.~Zadeh, M.~Chen, S.~Poria, E.~Cambria, and L.-P. Morency.
\newblock Tensor fusion network for multimodal sentiment analysis.
\newblock In {\em EMNLP}, 2017.

\bibitem{zadeh2018multimodal}
A.~B. Zadeh, P.~P. Liang, S.~Poria, E.~Cambria, and L.-P. Morency.
\newblock Multimodal language analysis in the wild: Cmu-mosei dataset and
  interpretable dynamic fusion graph.
\newblock In {\em Proceedings of the 56th Annual Meeting of the Association for
  Computational Linguistics (Volume 1: Long Papers)}, pages 2236--2246, 2018.

\end{thebibliography}
}

\appendix

\section{Supplementary Material}



This supplementary material provides more insights into our proposed methodology. In the section-A.1, we elaborate how KD is applied at the transformer level. In section-A.2 we visualize KD in action and finally in section-A.3 we disclose the training details of our experiments.

\subsection{KD at transformer level}

In section 4.3, we proposed two levels at which KD can be applied inside the transformer architecture, KD at attention map and KD at post-attention linear layers. 

\begin{figure*}
    \includegraphics[width=\textwidth]{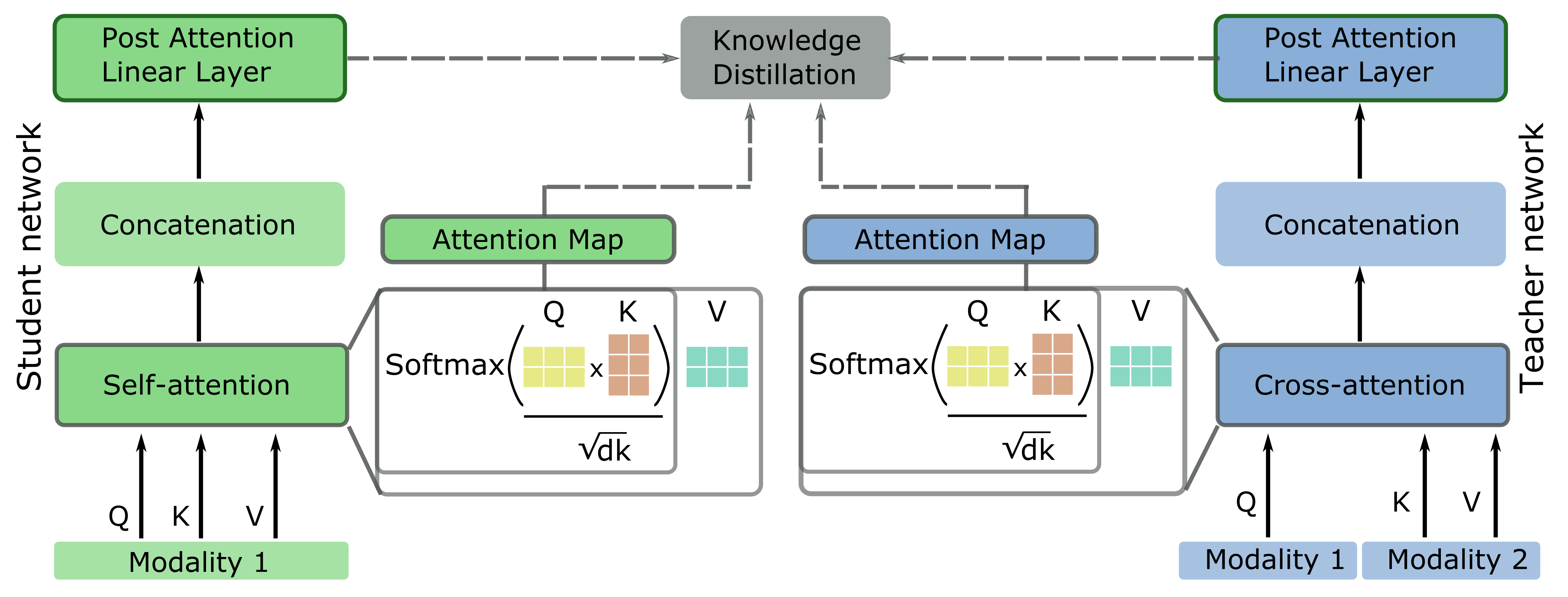}
    \caption{ \textbf{Knowledge distillation in the student-teacher transformers.} Representation of the two deeper layers, in the transformers architecture, where distillation is applied, at attention maps and at the post-attention linear layer.}
    \label{fig:KD_image}
\end{figure*}

In the attention module of a transformer, Q, K, and V $\in$  $R^{n \times d}$ (where n = sequence length, and d = feature embedding length) are are divided into n\textsubscript{h} chunks where n\textsubscript{h} is the number of attention heads in a transformer. The n\textsubscript{h} chunks of Q, K, and V $\in$ $R^{n \times d/n_h}$ undergo the attention mechanism (equation highlighted in the Fig-\ref{fig:KD_image}) to give $n_h$ outputs of attention maps. The average of all these $n_h$ attention maps is then serves as the attention map which is used for distillation

Similar to the attention maps, the output of the product of attention map with V gives a feature vector $\in$ $R^{n \times d/n_h}$. All these $n_h$ feature vectors are concatenated along the feature dimension to get back the feature vector $\in$ $R^{n \times d}$ (this operation is illustrated with \textbf{concatenation} layer in the Fig-\ref{fig:KD_image}). The resultant feature vector is then passed through linear layers to get what we call as post-attention linear layer, used for the distillation.

We extract these attention maps and post-attention linear layers for all the layers of the transformers and apply KD on all of them.

\label{sec:KD}

\subsection{Visualizing KD}

\begin{figure*}
 \centering
    \includegraphics{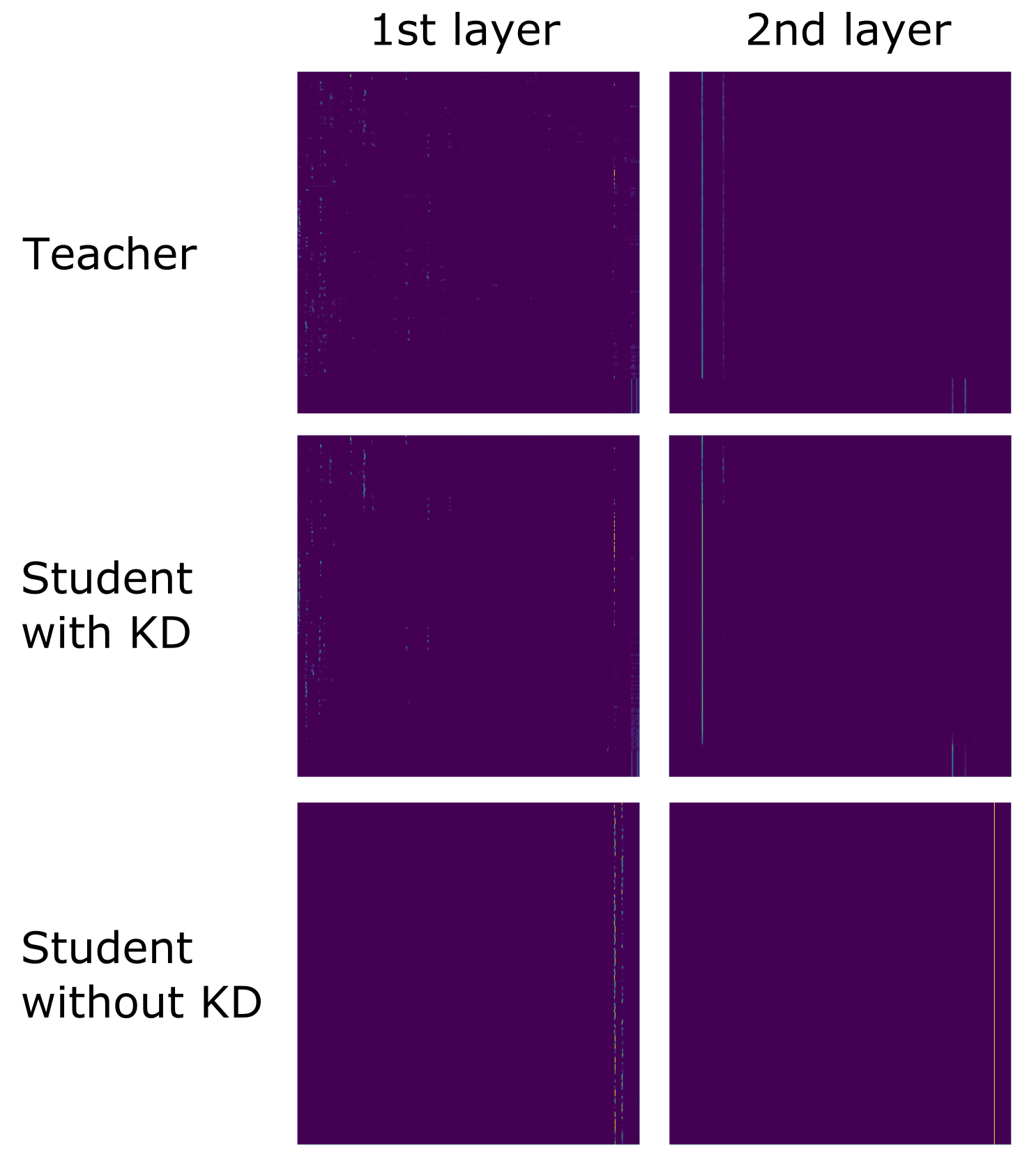}
    \caption{KD Visualization using EDAM-S$\downarrow$ on attention maps: Top-row = Attention maps from $1^{st}$ and $2^{nd}$ transformer layers of $V_T \leftarrow A_T$  respectively, Middle-row = Attention maps from $1^{st}$ and $2^{nd}$ layers of $V_S \leftarrow A_S$ after KD respectively, and bottom-row = Attention maps from $1^{st}$ and $2^{nd}$ layers of $V_S \leftarrow A_S$ before KD respectively}
    \label{fig:VizKD}
\end{figure*}

Here we present a way to visualize the effects of Knowledge Distillation on Attention Maps of the student transformers. In Fig-\ref{fig:VizKD}, we plot the attention maps of student and teacher transformers before and after applying EDAM-S$\downarrow$ on attention maps. From the figure we can clearly see that before distillation the attention maps of teacher network and student network are very different and thus both the models learn disparate representations. However, after applying EDAM-S$\downarrow$ on attention maps, the attention maps of student network begin to align themselves to that of teacher network, thus getting closer to the teacher network at lower levels. This we believe is conducive for learning richer representations, closer to teacher network, as the outputs from the attention maps are further passed to high-level layers which learn representations built upon low-level features. In Fig-\ref{fig:VizKD} we only provide the attention maps of one of the transformer pair due to lack of space, however, with EDAM-S$\downarrow$ on attention maps, all of the transformers from student closely resembled the corresponding teacher transformers.

\subsection{Training Details}

All of our experiments were performed on 1 Nvidia RTX-8000 with a batch size of 64. We used the same data splits (train, val, test) as used by \cite{tsai2019multimodal} in their approach. We trained our models on training set for 100 epochs and used the weights which gave lowest validation loss to compute results on test set. 

We used a learning rate of 1e-3 and decayed it with $ReduceLROnPlateau$ with a patience value of 10 epochs and a factor of 0.5. In all our transformers we had 4 layers with 8 attention heads.


\rule{0pt}{1pt}\newpage

\end{document}